\documentclass[conference]{IEEEtran}
\IEEEoverridecommandlockouts
\usepackage{cite}
\usepackage{amsmath,amssymb,amsfonts}
\usepackage{algorithmic}
\usepackage{graphicx}
\usepackage{textcomp}
\usepackage{booktabs}
\usepackage{multirow}
\usepackage{makecell}
\usepackage{xcolor}
\usepackage{array}
\usepackage{url}
\usepackage{booktabs,multirow}
\usepackage{algorithm}
\usepackage{soul}

\usepackage{hyperref}

\usepackage[dvipsnames]{xcolor}

\usepackage{subcaption}
\usepackage{tikz}
\newcommand{\legendline}[1]{%
  \tikz[baseline=-0.5ex]{%
    \draw[#1,line width=1pt] (0,0) -- (0.45,0);
  }%
}
\definecolor{LSTMcolor}{HTML}{FF6B6B}
\definecolor{NODEcolor}{HTML}{2EC4B6}
\definecolor{CfCcolor}{HTML}{4D96FF}
\definecolor{LTCcolor}{HTML}{9B5DE5}

\begin{document}

\title{LiquidMPPI: Liquid Neural Networks for Model-Based Control of Aerial Robots
\thanks{This work was supported in part by the Higher Education and Science Committee of the RA under Grant 24FP-C017.}}

\author{Amaras Nazarians \and Aren Karapetyan \and Astghik Hakobyan\thanks{
A. Nazarians is with the Center for Scientific Innovation and Education, Yerevan, Armenia
        ({\tt\small amaras\_nazarians26@alumni.aua.am}). 
        A. Karapetyan is with the Center for Scientific Innovation and Education, Yerevan, Armenia and Belimo Automation AG, Zürich, Switzerland ({\tt\small aren.karapetyan@alumni.ethz.ch}).
A. Hakobyan is with the American University of Armenia and National Polytechnic University of Armenia, Yerevan, Armenia
       ({\tt\small astghik.hakobyan@aua.am}).  
}
}

\maketitle

\begin{abstract}
Sampling-based quadrotor control requires repeated
dynamics predictions, motivating compact models that balance
predictive accuracy, computational efficiency, and tracking
performance. We present LiquidMPPI, which embeds liquid
neural dynamics within model predictive path integral
(MPPI) control. Training combines structured
excitation, MPPI-generated trajectories, and controller-in-the-loop data aggregation. We compare liquid time-constant (LTC) and
closed-form continuous-time (CfC) networks with long short-term
memory (LSTM) networks, neural ordinary differential equation
models, and an analytical baseline, without deployment-specific
retraining. CfC achieves the lowest mean aggressive-sine
prediction errors across three hidden dimensions, while
compact CfC and LTC support effective software-in-the-loop
tracking under a shared controller. In hardware demonstrations,
an eight-unit CfC predictor tracks all five nominal trajectories
and achieves approximately 19\% lower sine-tracking position
root-mean-square error than LSTM. These results highlight
liquid networks as effective compact rollout models and
the importance of jointly evaluating prediction accuracy
and closed-loop performance.
\end{abstract}


\section{Introduction}
\label{sec:introduction}

Predictive quadrotor control requires more than accurate
one-step predictions: a dynamics model must support
effective control decisions across candidate future maneuvers.
Model predictive control (MPC) addresses this objective
through repeated finite-horizon optimization and has enabled
agile quadrotor flight~\cite{bangura2014real,sun2022comparative}.
Model predictive path integral (MPPI) control is a
sampling-based MPC method that evaluates control sequences
without requiring dynamics
derivatives~\cite{MPPI_origin,Williams2017Information},
making it attractive for quadrotor
control~\cite{mohamed2020model,minavrik2024model}.
Evaluating thousands of trajectories per update motivates
compact rollout models that balance predictive accuracy,
computational cost, and closed-loop tracking performance.

Learned dynamics have expanded predictive quadrotor control
through Gaussian-process corrections~\cite{9361343},
neural prediction models~\cite{jiang2022neural},
physics-inspired temporal networks~\cite{saviolo2022physics},
and real-time neural MPC~\cite{salzmann2023real}, among other methods.
Recurrent architectures incorporate temporal information:
long short-term memory (LSTM) networks~\cite{LSTM}
have been used to predict quadrotor states within MPC
for formation control~\cite{najafqolian2022formation}.
Neural ordinary differential equations (NODEs)~\cite{NODE}
provide a continuous-time modeling alternative.
KNODE-MPC combines such learned dynamics with
first-principles models for quadrotor predictive
control~\cite{KNODE_MPC}, with related work enabling
online adaptation~\cite{Jiahao2022Online}.

Within MPPI, prior work has investigated neural quadrotor
dynamics~\cite{li2024quadcopter}, meta-learned adaptation
with uncertainty-aware planning~\cite{kalaria2025agile},
and transformer-based control-sequence
initialization~\cite{Zinage2024Transformer}.
Against this background, we examine a focused question:
\emph{how effectively can compact recurrent architectures
predict quadrotor motion and support closed-loop tracking
within the same MPPI framework?}

\begin{figure}[t]
    \centering
    \includegraphics[width=\columnwidth]
    {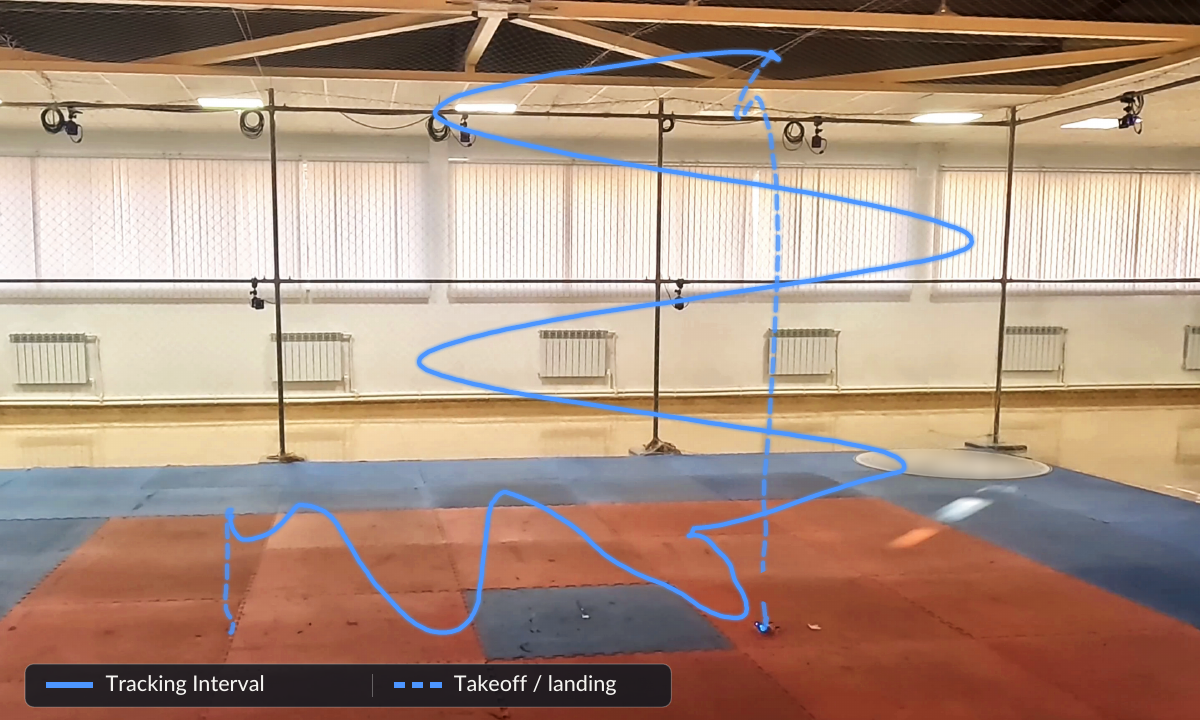}
\caption{Aggressive sine tracking on a physical Crazyflie
using LiquidMPPI with an eight-unit CfC predictor.
Solid and dashed paths indicate tracking and
takeoff/landing, respectively. See the \href{https://youtu.be/eT9covnyM94} {accompanying video}.}
    \label{fig:hardware_teaser}
\end{figure}

Liquid neural networks offer a promising architecture for
compact rollout modeling. Liquid time-constant (LTC)
networks use state- and input-dependent response
rates~\cite{LTC1}, whereas closed-form continuous-time
(CfC) networks provide explicit recurrent updates that
avoid numerical integration of ordinary differential
equations~\cite{CFC}.
Related architectures have supported compact autonomous
policies~\cite{NCP_auto}, vision-based flight
navigation~\cite{LTC_drone}, and disturbance learning
within quadrupedal MPC~\cite{mao2025liquid}.
Here, we investigate LTC and CfC as quadrotor rollout
models within MPPI: the networks predict motion under
candidate control sequences, while the controller
selects commands according to the tracking objective.

Evaluating these networks as rollout models requires
examining both prediction accuracy and closed-loop
performance: lower prediction error need not yield
better control~\cite{lambert2020objective}.
Control decisions influence the states and commands
encountered during execution, while feedback repeatedly
updates the predicted plan. Our evaluation therefore
examines both outcomes under nominal tracking,
aggressive maneuvers, and deployment beyond the training
environment.

We present \emph{LiquidMPPI}, a framework that
learns the rollout dynamics while retaining the MPPI
optimization procedure. Training combines pseudo-random
binary sequence (PRBS) excitation, MPPI-generated trajectories, and controller-in-the-loop data aggregation,
building on iterative collection principles in model-based
learning~\cite{nagabandi2018neural}.
The predictor learns velocity and attitude increments,
while position is updated using known kinematics. All training data are generated using the analytical model.
The trained predictors are then evaluated without retraining
in the analytical environment, CrazySim software-in-the-loop
(SITL), which couples Gazebo physics with Crazyflie
firmware~\cite{llanes2024crazysim}, and hardware demonstrations
(Fig.~\ref{fig:hardware_teaser}).

Our contributions are threefold:
\begin{enumerate}
    \item \emph{Liquid rollout models for MPPI.}
    We integrate LTC and CfC dynamics predictors through
    a reduced-state representation and controller-in-the-loop
    data aggregation, preserving the standard MPPI control
    interface and optimization procedure.

    \item \emph{Joint evaluation of prediction and control.}
We compare LTC and CfC with LSTM networks and recurrent
NODE models across hidden dimensions, alongside an
analytical baseline. The evaluation emphasizes compact
predictors and aggressive tracking under a shared
MPPI configuration.

    \item \emph{Hardware deployment without retraining.}
We demonstrate an eight-unit CfC predictor on a physical
Crazyflie across five nominal trajectories and an
aggressive sine maneuver, including a tracking comparison
with LSTM under shared controller settings.
\end{enumerate}

The remainder of this paper is organized as follows.
Section~\ref{sec:problem_formulation} introduces the
problem setup, MPPI formulation, and recurrent architectures.
Section~\ref{sec:liquidmppi} presents LiquidMPPI.
Section~\ref{sec:experiments} reports the experimental
evaluation, and Section~\ref{sec:conclusion} concludes
with future directions.


\section{Preliminaries}\label{sec:problem_formulation}

\subsection{Problem Setup}
\label{sec:quad_dynamics}

We consider the Crazyflie quadrotor with a cascaded control architecture: an external controller computes desired attitude and collective-thrust commands, while an onboard low-level controller stabilizes the attitude. We represent the outer-loop dynamics using a reduced-order analytical model with translational dynamics and a first-order closed-loop attitude response, without explicitly modeling individual motors or rotational rigid-body dynamics.

To this end, we define the quadrotor state as $    \mathbf{x}
    =
    [\mathbf{p}^{\top},\mathbf{v}^{\top},
    \boldsymbol{\eta}^{\top}]^{\top}
    \in \mathbb{R}^{9}$,
where $\mathbf{p}=[x,y,z]^{\top}$ and
$\mathbf{v}=[\dot{x},\dot{y},\dot{z}]^{\top}$ are world-frame
position and velocity, and
$\boldsymbol{\eta}=[\phi,\theta,\psi]^{\top}$ contains
roll, pitch, and yaw. The control input is $\mathbf{u}
    =
    [\boldsymbol{\eta}_{\mathrm{cmd}}^{\top},\Delta T]^{\top}
    \in \mathbb{R}^{4}, T=mg+\Delta T$,
where $\boldsymbol{\eta}_{\mathrm{cmd}}$ is the commanded attitude, $m$ is the nominal vehicle mass, $T$ is the collective thrust and $g$ is gravitational acceleration. The input $\Delta T$ is relative to the nominal level-hover thrust $mg$.

Neglecting aerodynamic drag and taking thrust along the
positive body vertical axis, the translational dynamics are
\begin{equation*}
    \dot{\mathbf{p}}=\mathbf{v},
    \qquad
    \dot{\mathbf{v}}
    =\frac{T}{m}R(\boldsymbol{\eta})\mathbf{e}_3-g\mathbf{e}_3,
\end{equation*}
where $\mathbf{e}_3=[0,0,1]^\top$. The rotation matrix $R(\boldsymbol{\eta})\in SO(3)$ maps body-frame vectors to the world frame, whose vertical axis points upward. The closed-loop attitude response is approximated by
\begin{equation*}
    \dot{\boldsymbol{\eta}}
    =D_{\tau}^{-1}
    (\boldsymbol{\eta}_{\mathrm{cmd}}-\boldsymbol{\eta}),
\end{equation*}
where $D_{\tau}=\operatorname{diag}
(\tau_{\phi},\tau_{\theta},\tau_{\psi})$ contains the attitude response time constants. Forward-Euler discretization
with sampling period $\Delta t$ yields the analytical
transition model $F_{\mathrm{math}}$. This model generates
training transitions, provides the MPPI rollout baseline,
and serves as the analytical evaluation environment.

\textbf{Tracking objective:}
Given a reference state trajectory $\{\mathbf{r}_t\}$ and
state observations, compute commands
$\mathbf{u}_t\in\mathcal{U}$ that minimize the tracking error
and the control effort over a finite prediction horizon,
where $\mathcal{U}$ is the admissible input set.
We investigate how the rollout model affects prediction
accuracy and closed-loop tracking, particularly for compact
recurrent models and aggressive trajectories.

\subsection{Model Predictive Path Integral Control}
\label{sec:mbc_mppi}

MPPI evaluates sampled control sequences without requiring
dynamics derivatives, allowing analytical and learned rollout
models to share the same optimization procedure.

At control step $t$, the controller maintains a nominal
sequence $U_t=(\mathbf{u}_t,\ldots,\mathbf{u}_{t+H-1})$
with prediction horizon $H$. Predicted states satisfy
\begin{equation*}
    \hat{\mathbf{x}}_{t+i+1}
    =F(\hat{\mathbf{x}}_{t+i},\mathbf{u}_{t+i}),
    \qquad \hat{\mathbf{x}}_t=\mathbf{x}_t,
\end{equation*}
where $\mathbf{x}_t$ is the measured state and $F$ denotes either the analytical model $F_{\mathrm{math}}$ or a learned predictor $F_\theta$. History dependence for recurrent predictors is omitted here for clarity.

For the predicted trajectory $X_t=\{\hat{\mathbf{x}}_{t+1},\ldots,\hat{\mathbf{x}}_{t+H}\}$,
the tracking objective is
\begin{equation}\label{cost}
\begin{aligned}
J(X_t,U_t)
    =\sum_{i=0}^{H-1}\Big[
      \|\hat{\mathbf{x}}_{t+i+1}-\mathbf{r}_{t,i} & \|_Q^2
      +\|\mathbf{u}_{t+i}\|_R^2\Big]\\
    &+\|\hat{\mathbf{x}}_{t+H}-\mathbf{r}_{t,H}\|_{Q_f}^2.
\end{aligned}
\end{equation}
Here, $\mathbf{r}_{t,i}$ denotes the stage reference for
$i=0,\ldots,H-1$, and $\mathbf{r}_{t,H}$ the terminal
reference. We define $\|\mathbf{a}\|_W^2=\mathbf{a}^{\top}W\mathbf{a}$;
$Q\succeq0$, $R\succ0$, and $Q_f\succeq0$ weight running
tracking error, control effort, and terminal error, respectively.

At each update, MPPI generates $N$ candidate sequences
$V_t^n=(\mathbf{v}_t^n,\ldots,\mathbf{v}_{t+H-1}^n)$
by sampling
$\boldsymbol{\epsilon}_{t+i}^{n}\sim\mathcal N(0,\Sigma)$
and setting
$\mathbf{v}_{t+i}^{n}
=\operatorname{clip}_{\mathcal U}
(\mathbf{u}_{t+i}+\boldsymbol{\epsilon}_{t+i}^{n})$.
Here, $\Sigma\succ0$ is the sampling covariance, and
clipping enforces the input bounds. Each candidate is propagated through $F$ to obtain $X_t^n$. With effective perturbations
$\delta\mathbf{u}_{t+i}^{n}=\mathbf{v}_{t+i}^{n}-\mathbf{u}_{t+i}$,
the sampling score is
\begin{equation*}
    S^n=\tfrac{1}{2}J(X_t^n,V_t^n)
    +\lambda\sum_{i=0}^{H-1}
    \mathbf{u}_{t+i}^{\top}\Sigma^{-1}\delta\mathbf{u}_{t+i}^{n}.
\end{equation*}
The normalized weights and updated commands are
\begin{equation*}
\begin{aligned}
    w^n\!=\frac{\exp[-(S^n-\rho)/\lambda]}
    {\sum_{j=1}^{N}\exp[-(S^j-\rho)/\lambda]},\,
    \mathbf{u}_{t+i}^{*}\!=\mathbf{u}_{t+i}+\sum_{n=1}^{N}w^n\delta\mathbf{u}_{t+i}^{n},
\end{aligned}
\end{equation*}
where $\rho=\min_n S^n$ improves numerical stability.
The temperature $\lambda>0$ controls sensitivity to rollout
cost differences, with smaller values favoring lower-cost
candidates more strongly.

Only the first updated command is applied. The sequence is
then shifted and extended to initialize the next optimization
from a new state observation. LiquidMPPI retains this procedure
while replacing the analytical rollout model with learned
liquid dynamics, as detailed in Section~\ref{sec:liquidmppi}.

\subsection{Recurrent Dynamics Learning}
\label{sec:recurrent_dynamics}

We compare two liquid neural network architectures,
LTC and CfC, with two recurrent baselines, LSTM and
a recurrent NODE model, within the same MPPI framework.
All four predict velocity and attitude increments from
recent state--control history, excluding absolute position
from the network inputs and outputs. Position is updated
using known kinematics, yielding a common rollout
interface across architectures.

Let $\mathbf s_t=[\mathbf v_t^\top,\boldsymbol\eta_t^\top]^\top\in\mathbb R^6$ denote the reduced state and $\mathbf z_t=[\mathbf s_t^\top,\mathbf u_t^\top]^\top\in\mathbb R^{10}$ the state--control input. For a history of length $L$, the complete learned transition is
\begin{equation*}
    \hat{\mathbf x}_{t+1}=F_\theta(\mathbf x_t,\mathcal H_t),
    \qquad \mathcal H_t=(\mathbf z_{t-L+1},\ldots,\mathbf z_t).
\end{equation*}
Here, $\theta$ collects the learned parameters, and the
final history input includes the candidate control
$\mathbf u_t$. The history provides temporal context
from recent states and commands.

All architectures process the $L$ inputs sequentially
and map the final hidden state
$\mathbf h^{(L)}\in\mathbb R^{n_h}$, where $n_h$ is
the hidden dimension, to a predicted increment through
a linear readout:
$\widehat{\Delta\mathbf s}_{t+1}
    =W_{\mathrm{out}}\mathbf h^{(L)}+\mathbf b_{\mathrm{out}}$,
where $W_{\mathrm{out}}\in\mathbb R^{6\times n_h}$ and
$\mathbf b_{\mathrm{out}}\in\mathbb R^6$ are the readout weight
matrix and bias vector, respectively. They are trained jointly with the
recurrent parameters to predict the target increment
$\Delta\mathbf s_{t+1}=\mathbf s_{t+1}-\mathbf s_t$. The full state is reconstructed as
\begin{equation*}
    \hat{\mathbf p}_{t+1}=\mathbf p_t+\Delta t\,\mathbf v_t,
    \qquad
    \hat{\mathbf s}_{t+1}=\mathbf s_t+\widehat{\Delta\mathbf s}_{t+1}.
\end{equation*}
Input and output scaling are omitted from the notation. During MPPI rollouts, each candidate starts from the available observation history. At subsequent prediction steps, the window is shifted and extended using predicted reduced states and candidate controls.

\paragraph{Discrete-time recurrent baseline}
We use an LSTM as the discrete-time recurrent baseline~\cite{LSTM}. Its single-step update is applied $L$ times:
\begin{equation*}
    (\mathbf h^{(j)},\mathbf c^{(j)})
    =f_\theta^{\mathrm{LSTM}}
    (\mathbf h^{(j-1)},\mathbf c^{(j-1)},\mathbf z_{t-L+j}),
\end{equation*}
for $j=1,\ldots,L$, where $\mathbf h^{(j)},\mathbf c^{(j)}\in\mathbb R^{n_h}$ are the hidden and cell states. Gates regulate how information is retained and modified across the window. The final hidden state $\mathbf h^{(L)}$ supplies the common readout.

\paragraph{Neural ODE baseline}
The NODE baseline advances its hidden state by integrating a learned vector field~\cite{NODE}:
\begin{equation*}
    \frac{d\mathbf h(\xi)}{d\xi}
    =f_\theta^{\mathrm{NODE}}(\mathbf h(\xi),\mathbf z_{t-L+j}),
\end{equation*}
where $\xi$ denotes internal integration time and $f_\theta^{\mathrm{NODE}}:\mathbb R^{n_h}\times\mathbb R^{10}\to\mathbb R^{n_h}$. During interval $j=1,\ldots,L$, the input is held fixed while the hidden state evolves. The next interval uses the next history input and inherits the preceding
interval's final hidden state. Our vector field uses an affine transformation followed by a hyperbolic tangent and fixed-step fourth-order Runge--Kutta integration, requiring multiple vector-field evaluations per interval.

\paragraph{Liquid time-constant networks}
LTC networks introduce state- and input-dependent effective time constants~\cite{LTC1,LTC2}. A representative formulation is
\begin{equation*}
    \frac{d\mathbf h}{d\xi}
    =-[\boldsymbol\tau^{-1}+f_\theta^{\mathrm{LTC}}(\mathbf h,\mathbf z)]\odot\mathbf h
    +f_\theta^{\mathrm{LTC}}(\mathbf h,\mathbf z)\odot\mathbf A.
\end{equation*}
Here, $\mathbf z=\mathbf z_{t-L+j}$ during interval
$j=1,\ldots,L$, $\boldsymbol\tau\in\mathbb R_{>0}^{n_h}$
contains the baseline time constants,
$\mathbf A\in\mathbb R^{n_h}$ is a learned parameter
vector, and $f_\theta^{\mathrm{LTC}}$ is a nonnegative
modulation function. Products and reciprocals are element-wise. The effective time constants
$(\boldsymbol\tau^{-1}+f_\theta^{\mathrm{LTC}})^{-1}$
allow hidden components to respond at input- and
state-dependent rates. We use a conductance-based
implementation with a fixed number of numerical solver
substeps per recurrent interval, carrying the hidden
state between intervals.

\paragraph{Closed-form continuous-time networks}
CfC networks use an explicit, time-modulated recurrent update inspired by LTC dynamics~\cite{CFC}. For $j=1,\ldots,L$, the gated variant is
\begin{equation*}
\begin{aligned}
    \boldsymbol\alpha^{(j)}
    &=\sigma\bigl(\mathbf a_\theta(\mathbf h^{(j-1)},\mathbf z_{t-L+j})\delta+\mathbf b_\theta(\mathbf h^{(j-1)},\mathbf z_{t-L+j})\bigr),\\
    \mathbf h^{(j)}
    &=(1-\boldsymbol\alpha^{(j)})\odot\mathbf g_\theta(\mathbf h^{(j-1)},\mathbf z_{t-L+j})\\
    &\quad+\boldsymbol\alpha^{(j)}\odot\mathbf p_\theta(\mathbf h^{(j-1)},\mathbf z_{t-L+j}),
\end{aligned}
\end{equation*}
where $\delta$ is the recurrent interval,
$\mathbf a_\theta,\mathbf b_\theta,\mathbf g_\theta,
\mathbf p_\theta$ are learned mappings into
$\mathbb R^{n_h}$, and $\sigma$ is the element-wise sigmoid. Unlike the NODE and LTC implementations, CfC requires no numerical solver substeps.\footnote{We use uniformly sampled data and a normalized CfC interval $\delta=1$, distinct from the physical sampling period $\Delta t$.}


\section{LiquidMPPI Framework}
\label{sec:liquidmppi}

\begin{figure}[t]
\centering
\includegraphics[width=\columnwidth]{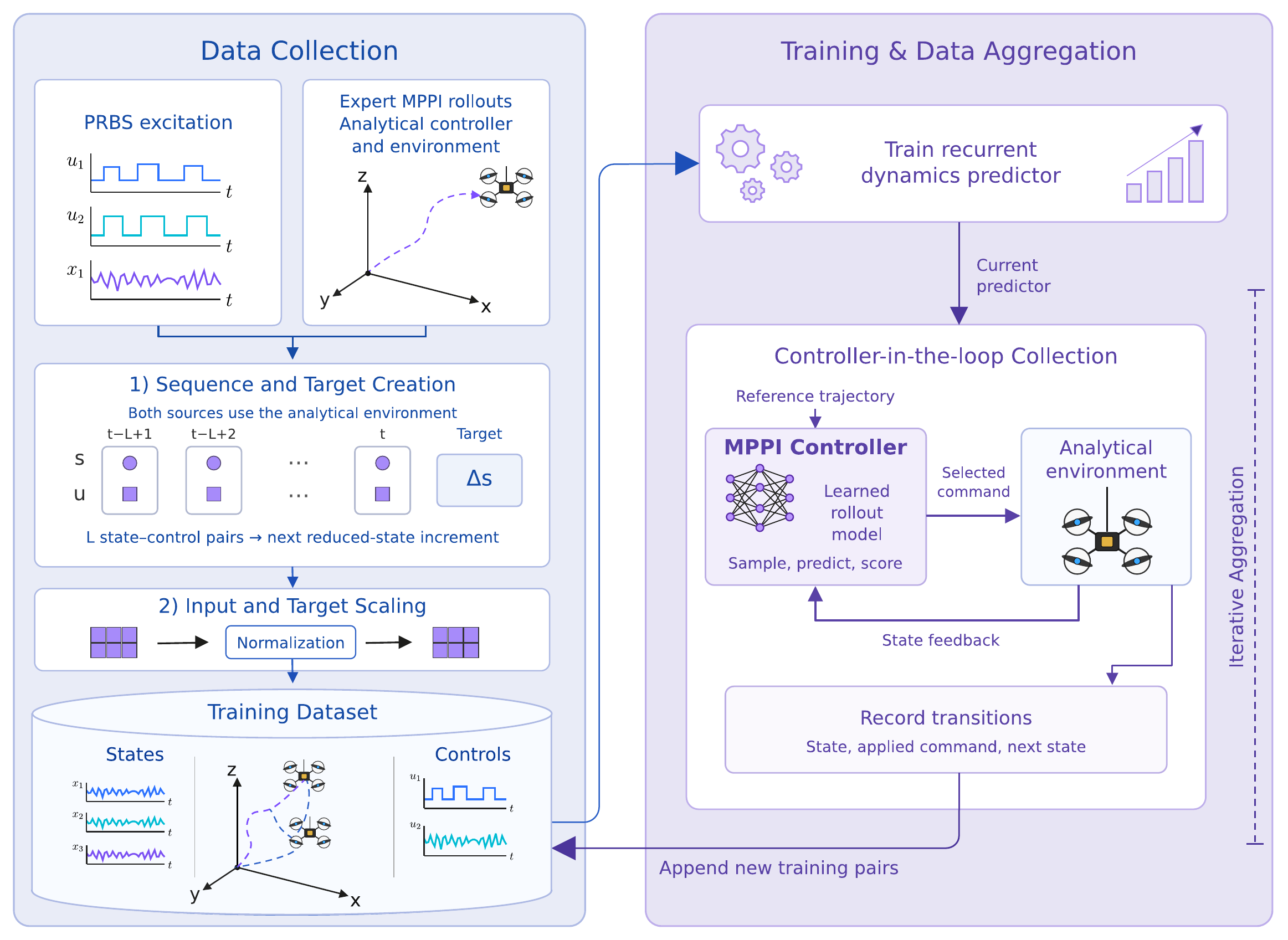}
\caption{Data collection and controller-in-the-loop training pipeline of LiquidMPPI.}
\label{fig:framework}
\end{figure}

LiquidMPPI couples dynamics learning with the controller
that uses its predictions. Its central idea is to
\textbf{train rollout models using both initial excitation
and tracking data, then expand the dataset through
learned-model MPPI}.
The pipeline combines data generation, recurrent prediction
with known position kinematics, and controller-in-the-loop
data aggregation. Additional transitions capture states
and commands encountered when the learned predictor guides
control. Figure~\ref{fig:framework} summarizes the framework.

\subsection{Rollout Integration}
\label{sec:learned_dynamics_representation}

The learned transition model $F_\theta$ defined in Section~\ref{sec:recurrent_dynamics} replaces the analytical model during MPPI prediction. At each control update, candidate rollouts share the same observed state and recorded past state--control history, but use different sampled control sequences.

Each candidate maintains its own history window, appending
predicted reduced states and candidate commands as the
horizon advances. Predictions are evaluated in batches,
and the reconstructed nine-dimensional states are scored
using the MPPI objective. After the selected command is
applied, the next optimization starts from a new state
observation rather than the previous prediction.

LTC, CfC, LSTM, and NODE share this rollout interface,
allowing comparisons without changing the state
representation or command interface.

\subsection{Data Collection}
\label{sec:expert_data_generation}

The initial dataset combines structured excitation and
trajectory-tracking data,
$\mathcal D^{(0)}
=\mathcal D_{\mathrm{PRBS}}\cup\mathcal D_{\mathrm{MPPI}}$,
generated in the analytical environment $F_{\mathrm{math}}$.
These components provide complementary coverage of responses
to changing commands and transitions encountered during
trajectory tracking.

\paragraph{Structured excitation}
We use PRBS, a standard excitation signal in system
identification~\cite{ljung1999system}, to probe responses
to changing attitude and thrust commands. Attitude commands
switch between prescribed levels, while thrust excitation
is expressed relative to nominal hover. These inputs
broaden coverage of command variations that may be
infrequent during nominal tracking.

\paragraph{MPPI tracking data}
For selected training references, MPPI uses $F_{\mathrm{math}}$
to select commands $\mathbf u_t^{*}$, which are
optionally perturbed before execution:
\begin{equation}
\label{eq:u_tilde_definition}
    \tilde{\mathbf u}_t
    =\mathbf u_t^*+\boldsymbol\nu_t,
    \qquad
    \boldsymbol\nu_t\sim\mathcal N(0,\Sigma_{\mathrm{aug}}),
\end{equation}
where $\Sigma_{\mathrm{aug}}$ is the collection-noise
covariance, distinct from the MPPI sampling covariance
$\Sigma$. The applied command is executed in the same
environment used for initial data generation.
Perturbations broaden the sampled transitions without
introducing additional physical effects.

Recorded trajectories are converted into history--target
pairs with target
$\Delta\mathbf s_{t+1}=\mathbf s_{t+1}-\mathbf s_t$.
Each input $\mathbf u_t$ in the training history
$\mathcal H_t$ is the command actually applied to the
environment, including any perturbation, rather than
necessarily the optimizer output $\mathbf u_t^*$.

\subsection{Controller-in-the-Loop Training}
\label{sec:training_pipeline}
The predictor is trained on history--increment pairs using
a squared-error objective in normalized target coordinates:
\begin{equation*}
    \mathcal L(\theta;\mathcal D)
    =\frac{1}{|\mathcal D|}
    \sum_{(\mathcal H,\Delta\mathbf s)\in\mathcal D}
    \Big\|
    \mathcal T_\Delta(\Delta\mathbf s)-\mathcal T_\Delta
    \bigl(\widehat{\Delta\mathbf s}_\theta(\mathcal H)\bigr)
    \Big\|_2^2,
\end{equation*}
where $\widehat{\Delta\mathbf s}_\theta(\mathcal H)$ is
the predicted increment in physical units and
$\mathcal T_\Delta$ is the target-scaling transformation.

Initial excitation and tracking data may not cover
the states and commands encountered when a learned
predictor guides MPPI. LiquidMPPI therefore periodically
embeds the current predictor in the controller and
collects additional transitions under its control decisions.

At aggregation iteration $k$, MPPI uses
$F_{\theta^{(k)}}$ for prediction, while the environment
generates the executed transitions:
\begin{equation*}
\begin{aligned}
    \mathbf u_t^*
    &=\operatorname{MPPI}\!\left(
    F_{\theta^{(k)}},\mathbf x_t,\mathcal H_t^-,
    \{\mathbf r_{t,i}\}_{i=0}^{H}\right),\\
    \mathbf x_{t+1}
    &=F_{\mathrm{math}}(\mathbf x_t,\tilde{\mathbf u}_t),
\end{aligned}
\end{equation*}
where $\tilde{\mathbf u}_t$ is defined in
\eqref{eq:u_tilde_definition} and
$\mathcal H_t^-=(\mathbf z_{t-L+1},\ldots,\mathbf z_{t-1})$
contains the preceding $L-1$ recorded state--control pairs.
Each candidate completes its input window by appending
the current reduced state and candidate command.


The dataset is expanded as
$\mathcal D^{(k+1)}
=\mathcal D^{(k)}\cup\mathcal D_{\mathrm{agg}}^{(k)}$,
and training continues on the accumulated data.
Thus, \emph{the learned predictor guides control decisions,
while the environment supplies the supervision}.

After training, the predictor and its scaling transformations
are frozen; evaluation uses neither online model adaptation
nor additional data aggregation.

\section{Experiments}
\label{sec:experiments}

We compare LTC and CfC liquid networks against LSTM
and recurrent NODE baselines across three hidden dimensions,
evaluating prediction accuracy and closed-loop tracking.
The evaluation emphasizes compact models and deployment
beyond the training environment. MPPI with the analytical
rollout model provides an additional control baseline
in the simulation comparisons.

We conduct five complementary evaluations:
\begin{enumerate}
    \item \textbf{Open-loop prediction:}
    autoregressive prediction under prescribed controls
    on nominal and aggressive sinusoidal trajectories,
    complemented by PRBS tests with increasing command
    amplitudes.

    \item \textbf{Analytical-environment tracking:}
    closed-loop evaluation of learned rollout models
    in the analytical environment.

    \item \textbf{Nominal SITL tracking:}
    evaluation in CrazySim~\cite{llanes2024crazysim},
    coupling Gazebo physics with Crazyflie firmware,
    under a fixed MPPI configuration shared across
    architectures and hidden dimensions.
    
    \item \textbf{Aggressive SITL tracking:}
    sinusoidal tracking in the same simulation stack
    under a separately selected MPPI configuration,
    also shared across architectures and hidden dimensions.

    \item \textbf{Hardware demonstrations:}
    nominal and aggressive sinusoidal tracking on a
    physical Crazyflie using compact CfC and LSTM
    predictors within MPPI.
\end{enumerate}

\subsection{Experimental Setup}
\label{sec:experimental_setup}

\paragraph{Models and training}
We consider hidden dimensions $n_h\in\{8,32,64\}$.
Initial datasets contain $2{,}000$ samples for $n_h=8$
and $20{,}000$ for $n_h\in\{32,64\}$, with equal
proportions of MPPI tracking and PRBS data.
Larger-model settings also collect ten times as many transitions from a broader trajectory set.
Thus, cross-dimension comparisons do not isolate model
capacity from training-data volume and coverage.

Training uses the normalized objective in
Section~\ref{sec:training_pipeline}, learning rate $0.01$,
batch size $32$, and at most $5{,}000$ epochs, stopping
after $2{,}000$ epochs without validation-loss improvement.
The training workstation has an NVIDIA RTX 5080 GPU
and Intel Core i9 CPU. Unless otherwise stated, we use history length $L=10$
in analytical-environment experiments and $L=4$ in SITL and hardware experiments, independently of the MPPI horizon $H$.

\paragraph{Fixed-controller tracking experiments}
The fixed configuration uses $H=10$, $N=10{,}000$,
$\lambda=3.3$, and sampling covariance
$\Sigma=\operatorname{diag}(0.1,0.1,0.01,0.01)$,
whose entries are variances before clipping.
The optimization weights are
$Q_{\mathrm{base}}=\operatorname{diag}(28.65,\allowbreak
20.17,\allowbreak 28.56,\allowbreak 25.64,\allowbreak
30.00,\allowbreak 1.40,\allowbreak 16.00,\allowbreak
20.53,\allowbreak 57.50)$
and
$R_{\mathrm{base}}=\operatorname{diag}(5.80,\allowbreak
8.72,\allowbreak 3.10,\allowbreak 0.01)$.
Obtained through L-SHADE tuning and manual
refinement~\cite{Tanabe2014LSHADE}, this configuration
is shared across architectures.
Reported costs follow the quadratic form in
\eqref{cost}, evaluated on executed states and applied
commands over the evaluation interval, using
$Q_{\mathrm{eval}}=Q_{\mathrm{base}}$ and
$R_{\mathrm{eval}}=R_{\mathrm{base}}$.
No terminal penalty is included in the reported costs.
Tracking and control costs denote the accumulated state-error
and input terms, respectively; total cost is their sum.

\paragraph{Shared-controller aggressive-sine experiment}
This SITL study uses $H=10$, $N=2{,}000$, and $L=4$.
A nine-candidate local sweep evaluates shared configurations
across architectures and dimensions using three validation
sampling seeds. Candidates are ranked by failure count,
then mean evaluation cost per step. The selected settings
are frozen before testing with separate sampling seeds.
This procedure is distinct from the L-SHADE tuning;
parameters and the failure criterion accompany the results.

\subsection{Results}
\label{sec:results}

\subsubsection{Open-Loop Dynamics Prediction}
\label{sec:open_loop_prediction}
We evaluate learned-model prediction accuracy using control
sequences generated by MPPI with the analytical rollout model
for tracking the nominal and aggressive sine trajectories. Each predictor
starts from the same initial state
and receives the same controls, but propagates its own predicted
states without measurement correction. Predictions are compared
with the resulting analytical-environment states, measuring
accumulated rollout error rather than one-step accuracy.

The nominal evaluation includes tilted square, horizontal
spiral, vertical figure-eight, horizontal square, and tilted
circle trajectories. An aggressive sinusoidal trajectory provides a separate stress test.
For each rollout, we compute accumulated Euclidean errors:
$E_{\mathrm{pos}}=\sum_{t=1}^{T_{\mathrm{eval}}-1}
\|\hat{\mathbf p}_t-\mathbf p_t\|_2,
E_{\mathrm{vel}}=\sum_{t=1}^{T_{\mathrm{eval}}-1}
\|\hat{\mathbf v}_t-\mathbf v_t\|_2,
E_{\mathrm{att}}=\sum_{t=1}^{T_{\mathrm{eval}}-1}
\|\hat{\boldsymbol\eta}_t-\boldsymbol\eta_t\|_2$.
Each rollout contains $T_{\mathrm{eval}}=1000$ state samples. Each checkpoint is evaluated in 100 trials
per trajectory. Nominal trial-mean errors are averaged
equally over the five trajectories within each checkpoint.

\begin{figure}[t]
\centering
\includegraphics[width=\linewidth]{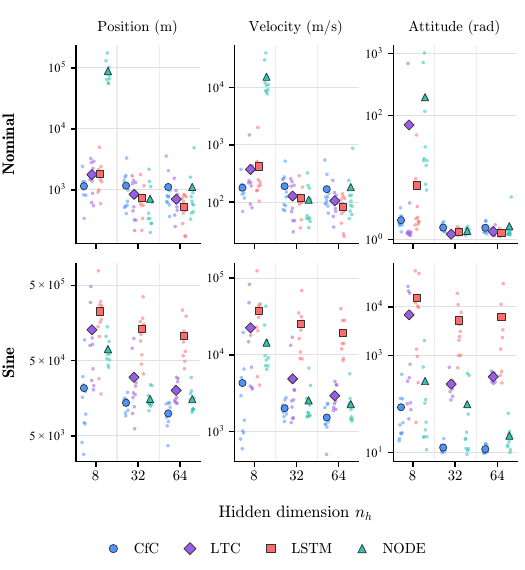}
\caption{Accumulated Euclidean prediction errors on nominal
(top) and aggressive sine (bottom) trajectories.
Dots show ten checkpoint averages; large
markers show their mean. Each averages 100 trials per
trajectory, equally weighting the five nominal trajectories.\vspace{-0.1in}}
\label{fig:open_loop_comparison}
\end{figure}

\begin{figure*}[ht!]
\centering
\includegraphics[width=0.85\textwidth]{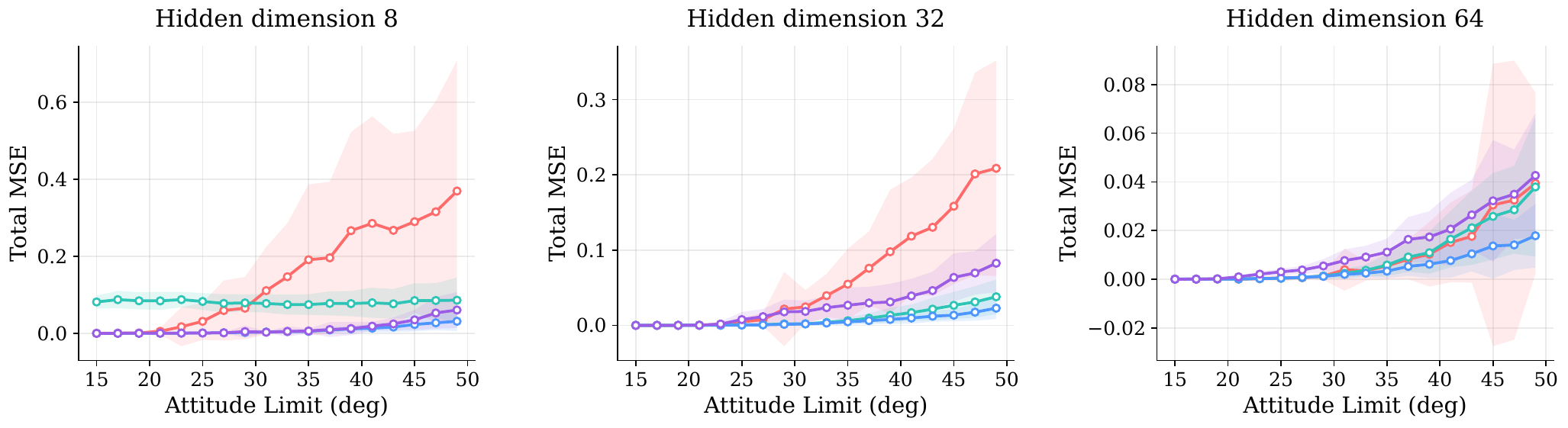}
\par\smallskip
{\footnotesize
\legendline{LSTMcolor}\ LSTM \quad
\legendline{NODEcolor}\ NODE \quad
\legendline{CfCcolor}\ CfC \quad
\legendline{LTCcolor}\ LTC\par}
\caption{Prediction error under increasing PRBS
attitude-command limits, using one selected checkpoint
per architecture and dimension and 100 rollouts of
50 steps per amplitude.}
\label{fig:prbs}
\end{figure*}

\paragraph{Nominal versus aggressive prediction}
Figure~\ref{fig:open_loop_comparison} summarizes ten
checkpoints per architecture and hidden dimension.
CfC performs particularly well in the compact setting:
at $n_h=8$, its mean nominal position, velocity, and
attitude errors are approximately $37\%$, $58\%$, and
$73\%$ lower than LSTM's, respectively. These results
demonstrate effective prediction with a small hidden
state and limited training data.

Under aggressive sine inputs, CfC achieves the lowest
mean error in every state group across all three
dimensions. Although NODE and LSTM become competitive
or lead on nominal trajectories at larger dimensions,
CfC retains its advantage on the aggressive trajectory.
The results therefore highlight both compact prediction
and performance under demanding motion as strengths
of CfC, while the checkpoint distributions reveal
variation across independently trained models.

\paragraph{Increased-amplitude excitation}
We further evaluate PRBS attitude-command limits from
$15^\circ$ to $50^\circ$, extending beyond the training
PRBS limit of $15^\circ$. Each amplitude uses 100
rollouts of 50 steps with one selected checkpoint per
architecture and hidden dimension.

Figure~\ref{fig:prbs} shows that compact CfC and LTC
maintain comparatively low errors as command amplitude
increases, whereas LSTM exhibits marked error growth.
This supports the ability of the evaluated compact
liquid models to predict responses to strong excitation.

Architectural differences narrow in the larger settings.
At $n_h=64$, the curves become closely grouped, although
CfC retains a modest advantage at the largest amplitudes.
These observations highlight the compact setting as
particularly informative for architectural comparison;
the larger settings combine increased capacity with
more extensive training data.

\subsubsection{Nominal Tracking in the Analytical Environment}
\label{sec:closed_loop_nominal}

We evaluate five checkpoints selected by open-loop performance
per architecture and dimension, using 100 trials on each
of the five nominal trajectories. Table~\ref{tab:closed_loop_id}
reports the means and sample standard deviations across
checkpoint-level costs, averaged over trials and equally
over trajectories.

\begin{table}[t!]
\centering
\caption{Nominal tracking in the analytical environment.
Costs are means $\pm$ sample standard deviations across
five selected checkpoints, each averaged over trials
and five trajectories.}
\label{tab:closed_loop_id}
\footnotesize
\setlength{\tabcolsep}{2pt}
\renewcommand{\arraystretch}{1.18}
\setlength{\thinmuskip}{1mu}

\begin{tabular*}{\columnwidth}{@{}@{\extracolsep{\fill}}clcc@{}}
\toprule
\textbf{Dim} & \textbf{Model}
& \textbf{Tracking Cost} & \textbf{Control Cost} \\
\midrule
\multirow{4}{*}{8}
 & LSTM & $917.30 \pm 76.82$  & $2.9496 \pm 0.5302$ \\
 & NODE & $1007.27 \pm 67.14$ & $2.7516 \pm 0.0275$ \\
 & CfC  & $887.08 \pm 18.93$
        & $\mathbf{2.7093 \pm 0.0240}$ \\
 & LTC  & $\mathbf{872.49 \pm 8.40}$
        & $2.7222 \pm 0.0090$ \\
\midrule
\multirow{4}{*}{32}
 & LSTM & $868.61 \pm 4.99$ & $2.7256 \pm 0.0179$ \\
 & NODE & $868.48 \pm 7.63$ & $2.7231 \pm 0.0216$ \\
 & CfC  & $\mathbf{865.94 \pm 5.32}$
        & $\mathbf{2.7191 \pm 0.0132}$ \\
 & LTC  & $868.78 \pm 2.11$ & $2.7269 \pm 0.0145$ \\
\midrule
\multirow{4}{*}{64}
 & LSTM & $978.54 \pm 172.50$ & $2.7311 \pm 0.0184$ \\
 & NODE & $916.28 \pm 81.98$  & $2.7204 \pm 0.0238$ \\
 & CfC  & $\mathbf{865.40 \pm 7.09}$
        & $\mathbf{2.7144 \pm 0.0119}$ \\
 & LTC  & $868.54 \pm 2.74$ & $2.7258 \pm 0.0130$ \\
\bottomrule
\end{tabular*}

\end{table}

All evaluated configurations complete the nominal tasks
without reported failures. Compact liquid models combine
favorable tracking and control costs: LTC achieves the
lowest tracking cost and smallest between-checkpoint
tracking variation, while CfC achieves the lowest control
cost. Their tracking performance therefore does not come
at the expense of increased control effort.

At $n_h=32$, the closely grouped costs indicate comparable
nominal performance across architectures. Increasing the
dimension further does not consistently improve tracking:
LSTM and NODE exhibit higher mean tracking costs and
greater checkpoint variability, whereas CfC and LTC
retain low costs and comparatively consistent performance.

The prediction and control rankings also differ.
LSTM's leading nominal open-loop performance at $n_h=64$
does not translate into the lowest tracking cost.
Unlike long autoregressive prediction, MPPI evaluates
candidate controls over a short horizon and repeatedly
replans from observed states providing closed-loop control. These results highlight
CfC and LTC as effective rollout models and reinforce
the importance of assessing learned dynamics within
the controller.

\subsubsection{Nominal Tracking in SITL}
\label{sec:gazebo}

We deploy the trained predictors in CrazySim SITL without
retraining, using the fixed MPPI configuration.
Each architecture and dimension uses one selected checkpoint,
evaluated in 100 trials per nominal trajectory with
$L=4$ and a nominal command rate of 50~Hz.

\begin{figure}
    \centering
    \includegraphics[width=\linewidth]{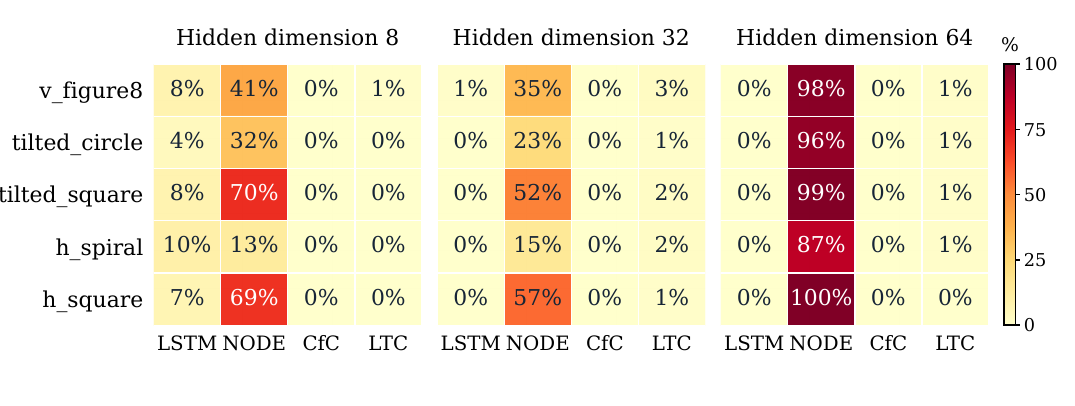}%
    \caption{CrazySim SITL failure rates over 100 trials per
model, dimension, and trajectory, using one selected
checkpoint.}
\label{fig:gazebo_fail}
\end{figure}

Figure~\ref{fig:gazebo_fail} reports failure rates.\footnote{Failure denotes minimum altitude $\leq0.08$~m during evaluation or nonfinite cost or altitude.}
Table~\ref{tab:gazebo_seq4_costs} summarizes pooled
successful, completed trials; trajectories therefore
contribute in proportion to their successful-run counts. CfC completes all nominal trials at every tested dimension,
supporting \textbf{deployment beyond the training environment
without model adaptation}. This consistent completion is
accompanied by favorable costs: among successful flights,
CfC achieves the lowest mean control cost at $n_h=32$
and $64$, and the lowest mean tracking cost at $n_h=32$.

LTC is particularly effective at $n_h=8$, where it
achieves the lowest successful-flight tracking and control
costs. At $n_h=64$, LTC again has the lowest tracking
cost, while CfC requires less control effort.
Together, these results highlight the practical value
of compact liquid rollout models.


\begin{table}[t]
\centering
\caption{Nominal tracking in CrazySim SITL.
Costs are means $\pm$ sample standard deviations over
successful trials pooled across five trajectories.}
\label{tab:gazebo_seq4_costs}

\footnotesize
\setlength{\tabcolsep}{2pt}
\renewcommand{\arraystretch}{1.18}
\setlength{\thinmuskip}{1mu}

\begin{tabular*}{\columnwidth}{@{}@{\extracolsep{\fill}}clcc@{}}
\toprule
\textbf{Dim} & \textbf{Model}
& \textbf{Tracking Cost} $(\times 10^3)$
& \textbf{Control Cost} \\
\midrule

\multirow{4}{*}{8}
& LSTM & $92.74 \pm 98.07$
& $76.80 \pm 39.97$ \\
& NODE & $395.97 \pm 731.85$
& $187.96 \pm 155.12$ \\
& CfC & $74.40 \pm 36.54$
& $64.26 \pm 21.78$ \\
& LTC & $\mathbf{70.70 \pm 33.91}$
& $\mathbf{39.96 \pm 17.77}$ \\

\midrule
\multirow{4}{*}{32}
& LSTM & $98.08 \pm 40.93$
& $68.26 \pm 19.68$ \\
& NODE & $343.38 \pm 517.74$
& $191.10 \pm 116.25$ \\
& CfC & $\mathbf{50.42 \pm 19.06}$
& $\mathbf{16.46 \pm 5.42}$ \\
& LTC & $283.05 \pm 176.62$
& $133.31 \pm 55.67$ \\

\midrule
\multirow{4}{*}{64}
& LSTM & $86.23 \pm 28.79$
& $47.08 \pm 10.67$ \\
& NODE & $3563.25 \pm 4863.47$
& $471.84 \pm 131.94$ \\
& CfC & $58.65 \pm 20.44$
& $\mathbf{16.49 \pm 4.68}$ \\
& LTC & $\mathbf{52.58 \pm 32.83}$
& $25.39 \pm 9.59$ \\

\bottomrule
\end{tabular*}

\end{table}

\begin{figure*}[t]
\centering
\captionsetup[subfigure]{font=footnotesize,skip=2pt}

{\footnotesize
\legendline{black,dashed}\ Reference \quad
\legendline{LSTMcolor}\ LSTM \quad
\legendline{NODEcolor}\ NODE \quad
\legendline{CfCcolor}\ CfC \quad
\legendline{LTCcolor}\ LTC
\par}

\begin{subfigure}[t]{0.195\textwidth}
    \centering
    \includegraphics[width=\linewidth]
      {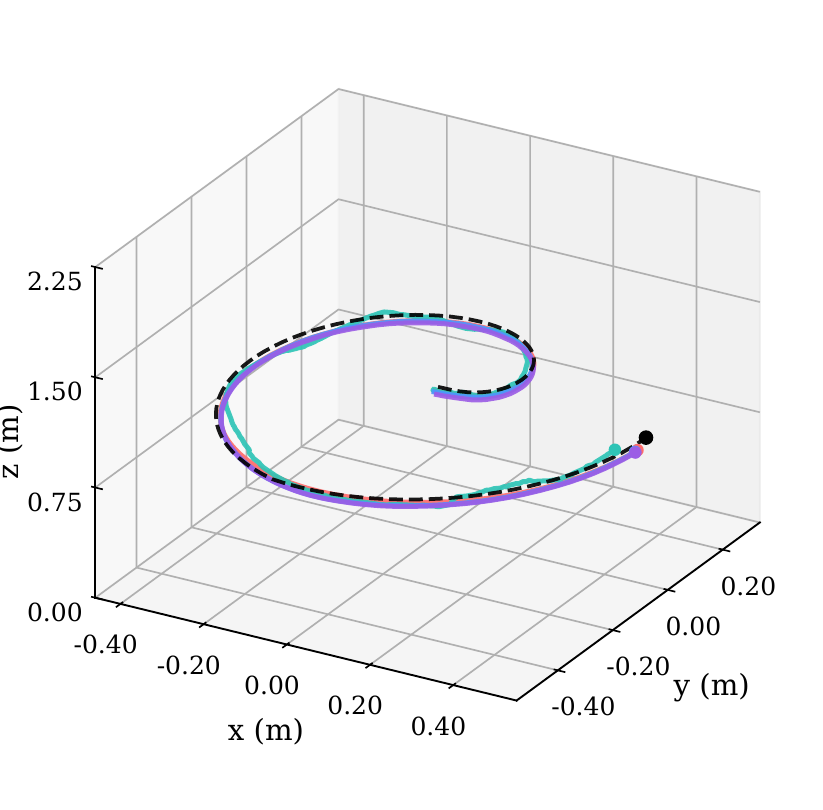}
    \caption{Horizontal spiral}
    \label{fig:gazebo_h_spiral}
\end{subfigure}\hfill%
\begin{subfigure}[t]{0.195\textwidth}
    \centering
    \includegraphics[width=\linewidth]
      {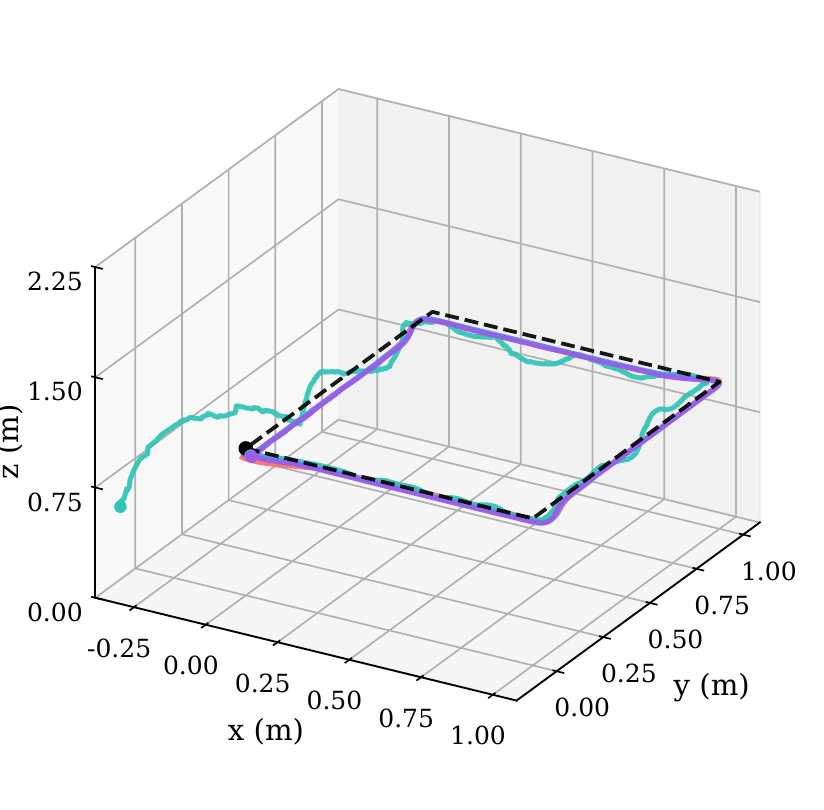}
    \caption{Horizontal square}
    \label{fig:gazebo_h_square}
\end{subfigure}\hfill%
\begin{subfigure}[t]{0.195\textwidth}
    \centering
    \includegraphics[width=\linewidth]
      {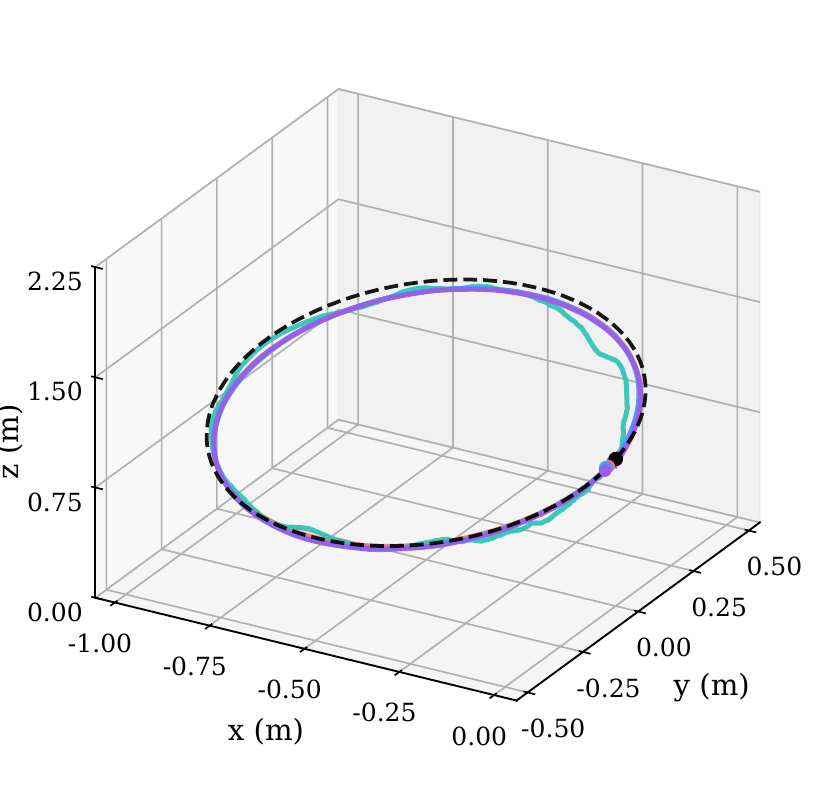}
    \caption{Tilted circle}
    \label{fig:gazebo_tilted_circle}
\end{subfigure}\hfill%
\begin{subfigure}[t]{0.195\textwidth}
    \centering
    \includegraphics[width=\linewidth]
      {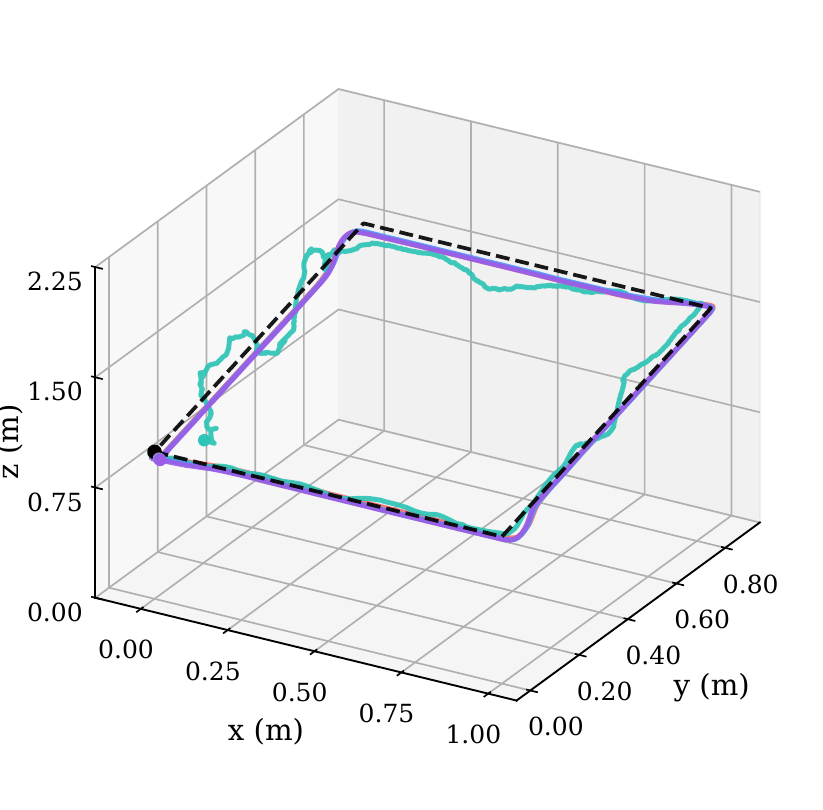}
    \caption{Tilted square}
    \label{fig:gazebo_tilted_square}
\end{subfigure}\hfill%
\begin{subfigure}[t]{0.195\textwidth}
    \centering
    \includegraphics[width=\linewidth]
      {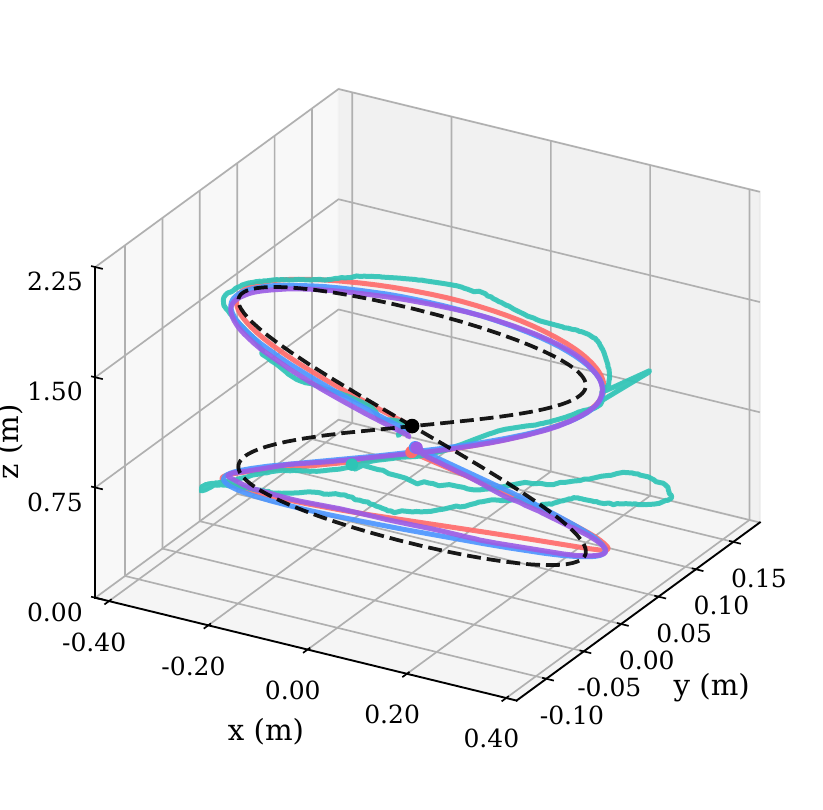}
    \caption{Vertical figure-eight}
    \label{fig:gazebo_v_figure8}
\end{subfigure}

\caption{Mean trajectories of successful CrazySim SITL
trials with $n_h=8$ and $L=4$.}
\label{fig:gazebo_five_trajectories}
\end{figure*}

Unlike the analytical environment, SITL reveals failures:
at $n_h=64$, NODE completes only 20 of 500 trials,
none on the horizontal square. Its costs thus reflect
a different trajectory mix, underscoring the need to
interpret successful-flight costs alongside completion rates.
Figure~\ref{fig:gazebo_five_trajectories} shows mean paths
from successful compact-model trials.

\subsubsection{Aggressive Sine Tracking in SITL}
\label{sec:gazebo_shared_sine}

We evaluate aggressive sine tracking in CrazySim SITL
under one MPPI configuration shared across architectures
and dimensions, including the analytical baseline.
The controller uses $H=10$, $N=2000$, $L=4$,
$\lambda=6.3$, and
$\Sigma=\operatorname{diag}(0.1,0.1,0.01,0.01)$.
Its weights are
$Q_{\mathrm{sin}}=\operatorname{diag}(2200,\allowbreak
2200,\allowbreak 9000,\allowbreak 300,\allowbreak
300,\allowbreak 450,\allowbreak 50,\allowbreak
50,\allowbreak 100)$ and
$R_{\mathrm{sin}}=\operatorname{diag}(12,18,5,2)$,
with reference starting position $(0,0,1)$~m.
Settings are selected using the validation procedure
in Section~\ref{sec:experimental_setup} and frozen
for 100 test flights per configuration with separate
sampling seeds.

\begin{table}[t]
\centering
\caption{Aggressive sine tracking in CrazySim SITL
under a shared controller. Cost and position RMSE are means $\pm$ sample standard deviations across successful trials.}
\label{tab:gazebo_shared_sine}
\footnotesize
\setlength{\tabcolsep}{2pt}
\renewcommand{\arraystretch}{1.05}
\begin{tabular*}{\columnwidth}{@{}@{\extracolsep{\fill}}clcc@{}}
\toprule
\textbf{Dim} & \textbf{Model}
& \textbf{Total Cost} $(\times 10^3)$
& \textbf{Position RMSE} (m) \\
\midrule
-- & Analytical & $3.933 \pm 0.154$ & $0.167 \pm 0.007$ \\
\midrule
\multirow{4}{*}{8}
& LSTM & $31.419$ & $0.819$ \\
& NODE & $4.701 \pm 0.227$ & $0.204 \pm 0.008$ \\
& CfC  & $4.434 \pm 0.285$ & $0.185 \pm 0.010$ \\
& LTC  & $4.405 \pm 0.309$ & $0.189 \pm 0.012$ \\
\midrule
\multirow{4}{*}{32}
& LSTM & $4.573 \pm 0.371$ & $0.200 \pm 0.015$ \\
& NODE & $4.286 \pm 0.432$ & $0.169 \pm 0.011$ \\
& CfC  & $4.362 \pm 0.261$ & $0.185 \pm 0.008$ \\
& LTC  & $4.980 \pm 1.449$ & $0.197 \pm 0.040$ \\
\midrule
\multirow{4}{*}{64}
& LSTM & $12.122 \pm 6.690$ & $0.401 \pm 0.172$ \\
& NODE & $5.046 \pm 0.923$ & $0.188 \pm 0.022$ \\
& CfC  & $4.510 \pm 0.249$ & $0.191 \pm 0.008$ \\
& LTC  & $4.963 \pm 0.297$ & $0.206 \pm 0.009$ \\
\bottomrule
\end{tabular*}
\end{table}

Table~\ref{tab:gazebo_shared_sine} summarizes only successful and
completed flights. Total cost is accumulated over
1,049 evaluated steps using the common evaluation weights;
failure counts are reported separately.
At $n_h=8$, CfC, LTC, and NODE complete all 100 flights, whereas LSTM fails in 99. The compact liquid models also achieve lower mean costs and position errors than NODE, demonstrating effective tracking without architecture-specific controller tuning.

Increasing dimension does not uniformly improve performance. At $n_h=32$, NODE achieves the lowest learned-model mean cost, and only LTC records a failure, with one failed flight. At $n_h=64$, CfC has the lowest learned-model mean cost; CfC, NODE, and LTC each fail once, compared with five failures for LSTM.

The analytical baseline completes every flight and achieves the lowest overall mean cost and position error. Among learned models, the results highlight strong compact performance from CfC and LTC and consistently competitive mean costs from CfC across dimensions.


\subsubsection{Hardware Demonstrations}
\label{sec:hardware}

We deploy compact CfC and LSTM predictors on a physical
Crazyflie using Vicon state feedback, without retraining.
Both use hidden dimension $n_h=8$ and history length
$L=4$, with identical MPPI settings within each comparison.

Under the shared nominal controller, CfC completes all
five nominal trajectories, whereas successful LSTM
tracking is observed only on the tilted square.
These demonstrations extend the evidence for compact
liquid rollout models to physical deployment.

For aggressive sine tracking, both models use a separately
tuned, shared MPPI configuration and complete the maneuver.
Position RMSE is $0.207$~m for CfC and $0.256$~m for LSTM,
giving CfC approximately $19\%$ lower error in the
demonstrated flights. Figure~\ref{fig:hardware_teaser}
shows the CfC flight. These results demonstrate deployment
feasibility and favorable CfC tracking, rather than
establishing statistical reliability.


\section{Conclusion and Future Work}
\label{sec:conclusion}

We presented LiquidMPPI, integrating liquid dynamics
learning and controller-in-the-loop data aggregation
within MPPI. CfC achieved the lowest mean aggressive-sine
prediction errors across tested dimensions,
while compact CfC and LTC enabled effective CrazySim
SITL tracking. In hardware demonstrations, CfC tracked
all five nominal trajectories and achieved approximately
$19\%$ lower sine-tracking position RMSE than LSTM.
These findings support liquid networks as effective
compact rollout models and highlight the importance
of evaluating both prediction and closed-loop performance.

Future work will explore learning from flight data,
systematic hardware comparisons, and evaluation under
matched computational budgets and external disturbances.

\bibliographystyle{IEEEtran}
\bibliography{ref} 

@article{salzmann2023real,
  author = {Salzmann, Tim and Kaufmann, Elia and Arrizabalaga, Jon and Pavone, Marco and Scaramuzza, Davide and Ryll, Markus},
  title = {Real-time neural {MPC}: Deep learning model predictive control for quadrotors and agile robotic platforms},
  journal = {IEEE Robot. Autom. Lett.},
  volume = {8},
  number = {4},
  pages = {2397--2404},
  year = {2023},
  doi = {10.1109/LRA.2023.3246839}
}

@inproceedings{najafqolian2022formation,
  author = {Najafqolian, Mohammad Amin and Alipour, Khalil and Mousavifard, Roujin and Tarvirdizadeh, Bahram},
  title = {Formation control of multiple aerial robots using {LSTM}-based model predictive control},
  booktitle = {Proc. RSI Int. Conf. Robot. Mechatronics},
  pages = {401--407},
  year = {2022}
}

@inproceedings{Tanabe2014LSHADE,
  author = {Tanabe, Ryoji and Fukunaga, Alex S.},
  title = {Improving the search performance of {SHADE} using linear population size reduction},
  booktitle = {Proc. IEEE Congr. Evol. Comput.},
  pages = {1658--1665},
  year = {2014},
  doi = {10.1109/CEC.2014.6900380}
}

@article{bangura2014real,
  author = {Bangura, Moses and Mahony, Robert},
  title = {Real-time model predictive control for quadrotors},
  journal = {IFAC Proc. Vol.},
  volume = {47},
  number = {3},
  pages = {11773--11780},
  year = {2014}
}

@article{sun2022comparative,
  author = {Sun, Sihao and Romero, Angel and Foehn, Philipp and Kaufmann, Elia and Scaramuzza, Davide},
  title = {A comparative study of nonlinear {MPC} and differential-flatness-based control for quadrotor agile flight},
  journal = {IEEE Trans. Robot.},
  volume = {38},
  number = {6},
  pages = {3357--3373},
  year = {2022}
}

@book{ljung1999system,
  author = {Ljung, Lennart},
  title = {System Identification: Theory for the User},
  publisher = {Prentice Hall},
  edition = {2},
  year = {1999}
}

@inproceedings{minavrik2024model,
  author = {Mina{\v{r}}{\'\i}k, Michal and P{\v{e}}ni{\v{c}}ka, Robert and Von{\'a}sek, Vojt{\v{e}}ch and Saska, Martin},
  title = {Model predictive path integral control for agile unmanned aerial vehicles},
  booktitle = {Proc. IEEE/RSJ Int. Conf. Intell. Robots Syst.},
  pages = {13144--13151},
  year = {2024}
}

@inproceedings{mohamed2020model,
  author = {Mohamed, Ihab S. and Allibert, Guillaume and Martinet, Philippe},
  title = {Model predictive path integral control framework for partially observable navigation: A quadrotor case study},
  booktitle = {Proc. Int. Conf. Control Autom. Robot. Vis.},
  pages = {196--203},
  year = {2020}
}

@article{jiang2022neural,
  author = {Jiang, Bailun and Li, Boyang and Zhou, Weifeng and Lo, Li-Yu and Chen, Chih-Keng and Wen, Chih-Yung},
  title = {Neural network based model predictive control for a quadrotor {UAV}},
  journal = {Aerospace},
  volume = {9},
  number = {8},
  pages = {460},
  year = {2022},
  doi = {10.3390/aerospace9080460}
}

@article{LSTM,
  author = {Hochreiter, Sepp and Schmidhuber, J{\"u}rgen},
  title = {Long short-term memory},
  journal = {Neural Comput.},
  volume = {9},
  number = {8},
  pages = {1735--1780},
  year = {1997},
  doi = {10.1162/neco.1997.9.8.1735}
}

@inproceedings{NODE,
  author = {Chen, Ricky T. Q. and Rubanova, Yulia and Bettencourt, Jesse and Duvenaud, David K.},
  title = {Neural ordinary differential equations},
  booktitle = {Adv. Neural Inf. Process. Syst.},
  volume = {31},
  year = {2018}
}

@inproceedings{LTC1,
  author = {Hasani, Ramin M. and Lechner, Mathias and Amini, Alexander and Rus, Daniela and Grosu, Radu},
  title = {Liquid time-constant networks},
  booktitle = {Proc. AAAI Conf. Artif. Intell.},
  volume = {35},
  pages = {7657--7666},
  year = {2021},
  doi = {10.1609/aaai.v35i9.16936}
}

@misc{LTC2,
  author = {Hasani, Ramin M. and Lechner, Mathias and Amini, Alexander and Rus, Daniela and Grosu, Radu},
  title = {Liquid time-constant recurrent neural networks as universal approximators},
  howpublished = {arXiv:1811.00321},
  year = {2018},
}

@article{LTC_drone,
  author = {Chahine, Makram and Hasani, Ramin M. and Kao, Patrick and Ray, Aaron and Shubert, Ryan and Lechner, Mathias and Amini, Alexander and Rus, Daniela},
  title = {Robust flight navigation out of distribution with liquid neural networks},
  journal = {Sci. Robot.},
  volume = {8},
  number = {77},
  pages = {eadc8892},
  year = {2023},
  doi = {10.1126/scirobotics.adc8892}
}

@article{NCP_auto,
  author = {Lechner, Mathias and Hasani, Ramin and Amini, Alexander and Henzinger, Thomas A. and Rus, Daniela and Grosu, Radu},
  title = {Neural circuit policies enabling auditable autonomy},
  journal = {Nat. Mach. Intell.},
  volume = {2},
  number = {10},
  pages = {642--652},
  year = {2020}
}

@article{CFC,
  author = {Hasani, Ramin M. and Lechner, Mathias and Amini, Alexander and Liebenwein, Lucas and Ray, Aaron and Tschaikowski, Max and Teschl, Gerald and Rus, Daniela},
  title = {Closed-form continuous-time neural networks},
  journal = {Nat. Mach. Intell.},
  volume = {4},
  number = {11},
  pages = {992--1003},
  year = {2022},
  doi = {10.1038/s42256-022-00556-7}
}

@misc{MPPI_origin,
  author = {Williams, Grady and Aldrich, Andrew and Theodorou, Evangelos},
  title = {Model predictive path integral control using covariance variable importance sampling},
  howpublished = {arXiv:1509.01149},
  year = {2015}
  }

@article{KNODE_MPC,
  author = {Chee, Kong Yao and Jiahao, Tom Z. and Hsieh, M. Ani},
  title = {{KNODE-MPC}: A knowledge-based data-driven predictive control framework for aerial robots},
  journal = {IEEE Robot. Autom. Lett.},
  volume = {7},
  number = {2},
  pages = {2819--2826},
  year = {2022},
  doi = {10.1109/LRA.2022.3144787}
}

@inproceedings{Jiahao2022Online,
  author = {Jiahao, Tom Z. and Chee, Kong Yao and Hsieh, M. Ani},
  title = {Online dynamics learning for predictive control with an application to aerial robots},
  booktitle = {Proc. Conf. Robot Learn.},
  series = {Proc. Mach. Learn. Res.},
  volume = {205},
  pages = {2251--2261},
  year = {2023}
}

@article{saviolo2022physics,
  author = {Saviolo, Alessandro and Li, Guanrui and Loianno, Giuseppe},
  title = {Physics-inspired temporal learning of quadrotor dynamics for accurate model predictive trajectory tracking},
  journal = {IEEE Robot. Autom. Lett.},
  volume = {7},
  number = {4},
  pages = {10256--10263},
  year = {2022}
}

@article{li2024quadcopter,
  author = {Li, Yong and Zhu, Qidan and Elahi, Ahsan},
  title = {Quadcopter trajectory tracking based on model predictive path integral control and neural network},
  journal = {Drones},
  volume = {9},
  number = {1},
  pages = {9},
  year = {2025},
  doi = {10.3390/drones9010009}
}

@inproceedings{kalaria2025agile,
  author = {Kalaria, Dvij and Xue, Haoru and Xiao, Wenli and Tao, Tony and Shi, Guanya and Dolan, John M.},
  title = {Agile mobility with rapid online adaptation via meta-learning and uncertainty-aware {MPPI}},
  booktitle = {Proc. IEEE Int. Conf. Robot. Autom.},
  pages = {16626--16632},
  year = {2025}
}

@misc{Zinage2024Transformer,
  author = {Zinage, Shrenik and Zinage, Vrushabh and Bakolas, Efstathios},
  title = {Transformer-based model predictive path integral control},
  howpublished = {arXiv:2412.17118},
  year = {2024}
}

@inproceedings{Williams2017Information,
  author = {Williams, Grady and Wagener, Nolan and Goldfain, Brian and Drews, Paul and Rehg, James M. and Boots, Byron and Theodorou, Evangelos A.},
  title = {Information-theoretic model predictive control for model-based reinforcement learning},
  booktitle = {Proc. IEEE Int. Conf. Robot. Autom.},
  pages = {1714--1721},
  year = {2017}
}

@article{9361343,
  author = {Torrente, Guillem and Kaufmann, Elia and F{\"o}hn, Philipp and Scaramuzza, Davide},
  title = {Data-driven {MPC} for quadrotors},
  journal = {IEEE Robot. Autom. Lett.},
  volume = {6},
  number = {2},
  pages = {3769--3776},
  year = {2021},
  doi = {10.1109/LRA.2021.3061307}
}

@inproceedings{llanes2024crazysim,
  author = {Llanes, Christian and Kakish, Zahi and Williams, Kyle and Coogan, Samuel},
  title = {{CrazySim}: A software-in-the-loop simulator for the {Crazyflie} nano quadrotor},
  booktitle = {Proc. IEEE Int. Conf. Robot. Autom.},
  pages = {12248--12254},
  year = {2024},
  doi = {10.1109/ICRA57147.2024.10610906}
}

@inproceedings{nagabandi2018neural,
  author = {Nagabandi, Anusha and Kahn, Gregory and Fearing, Ronald S. and Levine, Sergey},
  title = {Neural network dynamics for model-based deep reinforcement learning with model-free fine-tuning},
  booktitle = {Proc. IEEE Int. Conf. Robot. Autom.},
  pages = {7559--7566},
  year = {2018},
  doi = {10.1109/ICRA.2018.8463189}
}

@inproceedings{lambert2020objective,
  author = {Lambert, Nathan and Amos, Brandon and Yadan, Omry and Calandra, Roberto},
  title = {Objective mismatch in model-based reinforcement learning},
  booktitle = {Proc. Conf. Learn. Dyn. Control},
  series = {Proc. Mach. Learn. Res.},
  volume = {120},
  pages = {761--770},
  year = {2020}
}

@article{mao2025liquid,
  author = {Mao, Yuhui and Zhang, Yaxing and Gao, Longsen},
  title = {Liquid-augmented {MPC} in quadrupedal robot for disturbance learning},
  journal = {Electronics},
  volume = {14},
  number = {24},
  pages = {4843},
  year = {2025},
  doi = {10.3390/electronics14244843}
}

\end{document}